\documentclass[letterpaper, 10 pt, conference]{ieeeconf}  

\IEEEoverridecommandlockouts 

\usepackage{graphics} 
\usepackage{graphicx}
\usepackage{url}
\usepackage{booktabs}
\usepackage{multirow}
\usepackage{epsfig} 
\usepackage{subcaption}
\usepackage{amsmath} 
\usepackage{amssymb}  
\usepackage{amsfonts}
\usepackage{algorithm}
\usepackage{xcolor}
\usepackage{hyperref}
\usepackage{tabularx}

\title{\LARGE \bf
A Contrastive Few-shot RGB-D Traversability Segmentation Framework for Indoor Robotic Navigation} 

\author{Qiyuan An$^{1,2}$, Tuan Dang$^{3}$, and Fillia Makedon$^{1}$ 
\thanks{$^{1}$Qiyuan An was with Department of Computer Science and Engineering, University of Texas at Arlington, 1225 West Mitchell, Arlington, TX 76019, USA {\tt\small qxa5560@mavs.uta.edu}. He is now with Uber, Sunnyvale, CA 94086 USA {\tt\small qiyuan.an@uber.com}.}
\thanks{$^{3}$Tuan Dang is with Cognitive Robotics Lab, Department of Electrical Engineering and Computer Science, University of Arkansas, Fayetteville, AR, USA {\tt\small tuand@uark.edu}}
\thanks{$^{1}$Fillia Makedon is with Department of Computer Science and Engineering, University of Texas at Arlington, 1225 West Mitchell, Arlington, TX 76019, USA.  {\tt\small makedon@uta.edu}}
}

\begin{document}

\maketitle
\thispagestyle{empty}
\pagestyle{empty}

\begin{abstract}
Indoor traversability segmentation aims to identify safe, navigable free space for autonomous agents, which is critical for robotic navigation. Pure vision-based models often fail to detect thin obstacles, such as chair legs, which can pose serious safety risks. We propose a multi-modal segmentation framework that leverages RGB images and sparse 1D laser depth information to capture geometric interactions and improve the detection of challenging obstacles. To reduce the reliance on large labeled datasets, we adopt the few-shot segmentation (FSS) paradigm, enabling the model to generalize from limited annotated examples. Traditional FSS methods focus solely on positive prototypes, often leading to overfitting to the support set and poor generalization. To address this, we introduce a negative contrastive learning (NCL) branch that leverages negative prototypes (obstacles) to refine free-space predictions. Additionally, we design a two-stage attention depth module to align 1D depth vectors with RGB images both horizontally and vertically. Extensive experiments on our custom-collected indoor RGB-D traversability dataset demonstrate that our method outperforms state-of-the-art FSS and RGB-D segmentation baselines, achieving up to 9\% higher mIoU under both 1-shot and 5-shot settings. These results highlight the effectiveness of leveraging negative prototypes and sparse depth for robust and efficient traversability segmentation.
\end{abstract}

\section{Introduction}
Traversability segmentation is a fundamental task in robotic navigation, aiming to identify freespace that is safe for autonomous agents to traverse. While most existing work focuses on outdoor scenarios such as self-driving cars \cite{hosseinpoor2021traversability,sevastopoulos2022survey}, indoor environments remain relatively underexplored despite their practical importance in warehouse automation, hotel service, and hospital robotics \cite{an2024enhancing,an2024few}. Compared to outdoor scenes that often feature structured roads and lane markings, indoor traversability segmentation poses unique challenges due to varying lighting conditions, complex floor textures, cluttered layouts, and the presence of arbitrarily moving humans or objects \cite{sevastopoulos2023learning}.
\par
Most prior work adopts purely vision-based solutions \cite{hirose2018gonet,oh2022travel,watson2020footprints}. However, our analysis shows that even state-of-the-art segmentation models, such as Deeplabv3+ \cite{chen2017deeplab} and SegFormer \cite{xie2021segformer}, struggle to detect thin obstacles such as chair legs. Although these objects occupy only a small fraction of image pixels and have minimal influence on global segmentation metrics, failing to detect them can pose significant safety risks to robots.

\begin{figure}[ht!]
\centering
\includegraphics[width=1\linewidth]{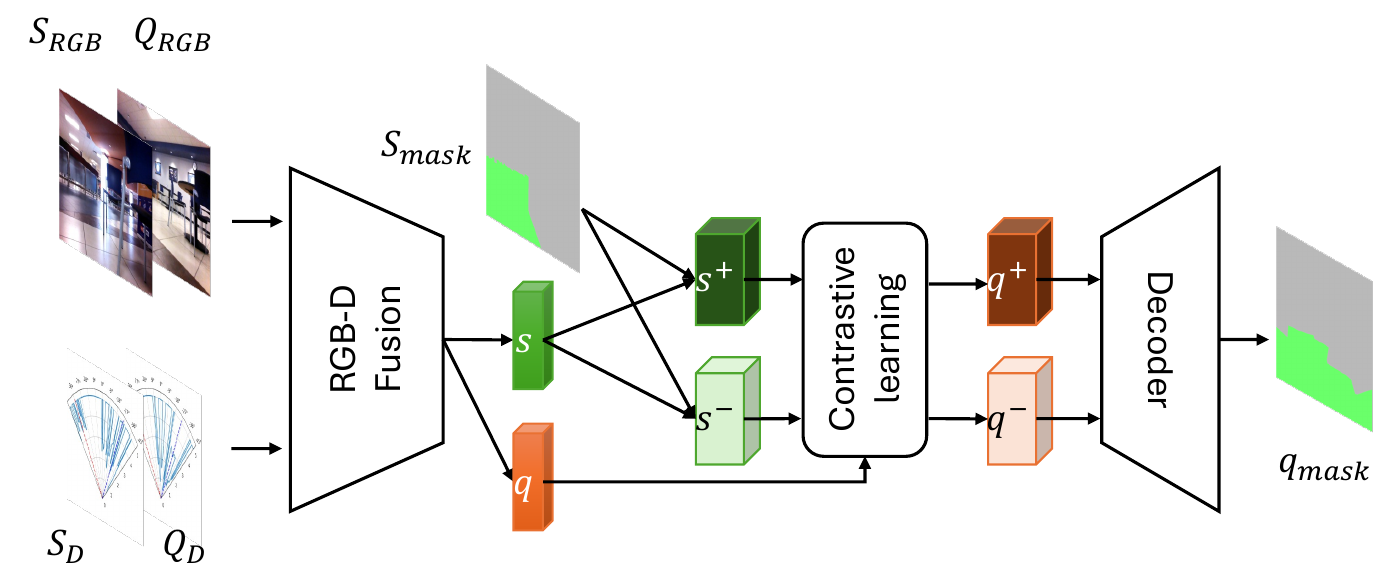}
\caption{Illustration of our RGB-D few-shot segmentation framework. The support and query inputs consist of RGB images ($S_{RGB}$, $Q_{RGB}$) and depth vectors ($S_{D}$, $Q_{D}$), encoded separately and fused via a multi-modal fusion block to produce support and query features ($s$, $q$). The support mask $S_{mask}$ is used to mask-pool on $s$, yielding both positive and negative prototypes ($s^+$, $s^-$). The query feature $q$ is then refined into free-space and obstacle representations ($q^+$, $q^-$), which are concatenated and digested by a lightweight decoder to generate the final query mask ($q_{mask}$).}
\label{fig:concept}
\end{figure}

\par
Unlike conventional RGB-D datasets such as NYUv2 \cite{silberman2012nyuv2} or SUN RGB-D \cite{song2015sun}, which provide dense 2D depth maps registered with corresponding RGB images, our custom-collected dataset consists of paired RGB images and 1D laser scans. We adopt this sensing setup for several reasons. First, many commercial indoor robots (e.g., cleaning, delivery, and assistive robots) are equipped with lightweight, low-cost 1D LiDARs rather than expensive 2D or 3D depth cameras, making our dataset more representative of real-world deployment scenarios. Second, 1D laser sensors enable large-scale, long-duration data collection at a fraction of the cost of dense depth sensors, thereby allowing us to build a large-scale dataset of thousands of RGB–1D depth pairs. Finally, the sparse and partial nature of 1D scans introduces a unique technical challenge: the depth backbone must learn to effectively encode 1D inputs and align them with 2D RGB features, which differs substantially from prior multimodal segmentation frameworks such as DFormer \cite{yin2023dformer} and CANet \cite{zhou2020rgbd}. 

\par
Therefore, our dataset is inherently more challenging than conventional RGB-D datasets: the 1D depth signals are vertically degenerated and often unregistered with respect to the RGB images. This setting reflects real-world conditions in which robot-mounted sensors are limited in resolution, field of view, or calibration accuracy. Consequently, models trained on our dataset must be robust to sensor imperfections, unregistration, and sparse depth information—properties that are critical for practical indoor robotic navigation.
\par
Another challenge comes from the unregistered depth vectors, which do not align well with the vertical beams of the corresponding RGB images in our custom-collected dataset. To address this, we introduce a two-stage attention depth module that dynamically maps the 1D depth to its paired RGB image along both vertical and horizontal dimensions. This design not only eliminates the need for explicit registration but also enables the model to capture dynamic geometric interactions between RGB and depth features.
\par 
Training a reliable traversability segmentation model also faces several practical challenges. Chief of them is acquiring large-scale, fine-grained annotations, which is often expensive, time-consuming, and labor-intensive \cite{casanova2020reinforced}. Few-shot segmentation (FSS) addresses this challenge by enabling models to learn from a limited number of labeled examples (the \textit{support} set) and generalize to new, unseen instances (the \textit{query} set) \cite{li2021adaptive,lu2021cwtfss,zhang2022mask}. We adopt the meta-learning paradigm in which the query set is matched to support prototypes at the feature level, allowing the model to generalize with minimal supervision.
\par
In the context of indoor traversability segmentation, traditional prototype-matching methods first learn a \textit{positive prototype} from the support set, representing traversable freespace. Query pixels that closely match this prototype in the feature space are then classified as freespace. However, this strategy is prone to misclassification when regions share similar textures or colors—for example, confusing white walls with white ceramic floor tiles. Such reliance on positive prototypes alone leads to overfitting to the support set and poor generalization to unseen scenarios.
\par
To address this limitation, we propose to explicitly leverage \textit{negative prototypes}—representations of obstacles that are typically ignored in conventional FSS methods. Our negative contrastive learning (NCL) branch first extracts obstacle prototypes from the support set and then identifies corresponding obstacle regions in the query set. These negative regions are used to refine the free-space masks by explicitly expelling obstacles from potential traversable areas. Finally, a lightweight decoder fuses the outputs of both the positive and negative branches to produce the final segmentation mask.
\par
To summarize, we propose a multi-modal traversability segmentation framework with a novel few-shot training paradigm. Our main contributions are:
\begin{enumerate}
\item Multi-modal RGB-D segmentation: Integrating RGB images and 1D depth vectors captures geometric interactions and improves detection of thin obstacles.  
\item Two-stage attention depth module: Dynamically aligning depth vectors with RGB images along horizontal and vertical dimensions addresses unregistered depth.  
\item Negative contrastive learning: Leveraging neglected negative prototypes enhances generalization and reduces overfitting in free-space prediction.  
\item Dataset contribution: We collect and release a large-scale indoor RGB-D traversability dataset with sparse 1D depth annotations, providing a new benchmark for future research in indoor navigation.
\end{enumerate}
\par
Our implementation will be publicly available at
\url{https://github.com/qiyuan53/NCL}.

\section{Related Work}
Traversability segmentation is a critical component of robotic navigation, as it enables autonomous agents to identify freespace while avoiding obstacles. Most existing approaches rely solely on visual information to extract features and recognize traversable regions \cite{hirose2018gonet,hirose2019vunet,hosseinpoor2021traversability,sevastopoulos2022indoorvit,an2024enhancing}. However, vision data alone often proves insufficient in complex environments with irregularly shaped objects such as appliances and furniture. For instance, even state-of-the-art segmentation models such as SegFormer \cite{xie2021segformer} struggles to exclude thin structures (e.g., chair legs) from freespaces, even under fully supervised training.
\par
To address these limitations, researchers have increasingly incorporated depth sensing modalities, including LiDAR \cite{cao2019adversarial,oh2022travel,zhu2021adversarial} and depth cameras \cite{yin2023dformer,yang2019robustifying}, to complement RGB information in traversability segmentation. While RGB–D fusion has demonstrated clear benefits, training reliable multi-modal segmentation models typically requires large quantities of fine-grained labeled data, which is costly and time-consuming to obtain.
\par
Few-shot learning (FSL) provides a promising alternative by enabling models to quickly adapt to new scenarios using only a limited number of labeled examples \cite{vinyals2016matching,snell2017prototypical}. Extending this paradigm, few-shot segmentation (FSS) adapts models on a small support set containing target classes and then infers masks on the query set. However, conventional FSS methods typically focus only on positive prototype matching—aligning query features with prototypes of the target class—while neglecting the background class \cite{shaban2017oneshot,wang2019panet,lu2021cwtfss,zhang2022mask}. In the context of traversability segmentation, this limitation is particularly problematic: positive prototype matching tends to bias the model toward a specific freespace type seen in the support set (e.g., carpet), while failing to generalize across diverse freespace appearances (e.g., ceramic tiles).
\par
To address this drawback, we introduce a novel negative contrastive learning branch that explicitly expels the background class (obstacles) to refine freespace predictions, improving generalization and robustness in unseen environments. To the best of our knowledge, this is the first work to explore few-shot RGB–1D depth traversability segmentation, bridging multi-modal fusion and FSS in a challenging real-world setting.

\begin{figure*}[t]
\centering
\includegraphics[width=0.99\linewidth]{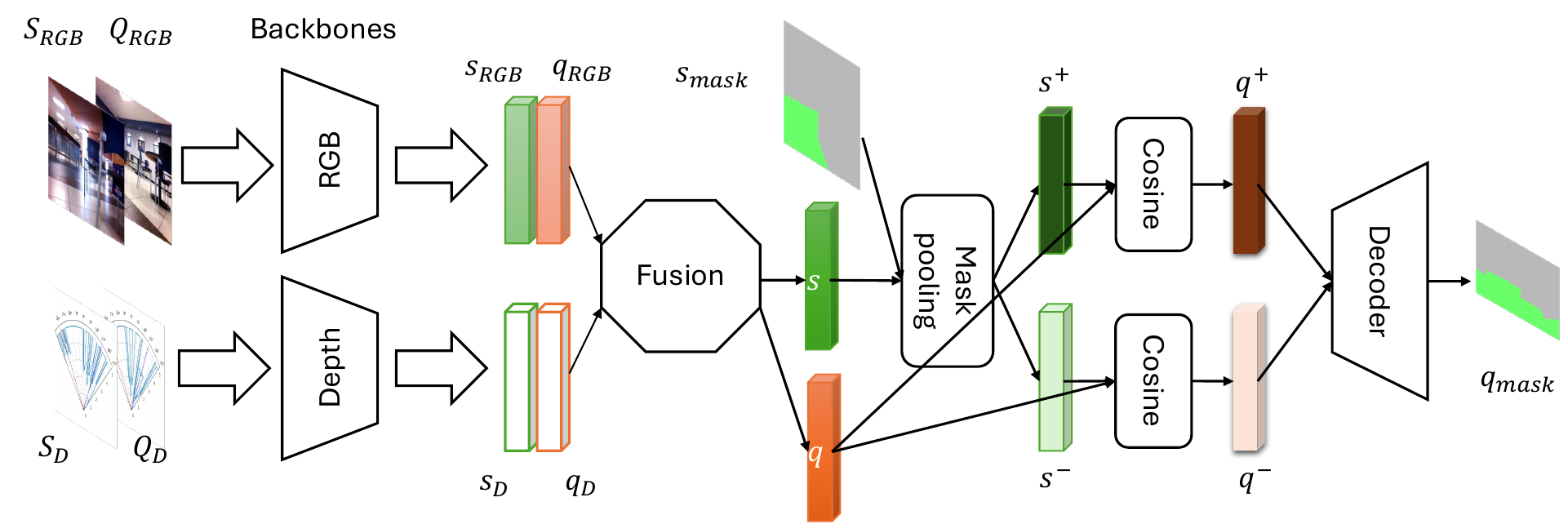}
\caption{Proposed contrastive few-shot RGB-D segmentation framework. RGB and depth inputs are embedded with modality-specific backbones, fused into unified support and query features, and refined through prototype-based contrastive learning. A lightweight decoder then predicts the query segmentation mask for indoor freespaces.}
\label{fig:main_arch}
\end{figure*}

\begin{figure}[ht]
\centering
\includegraphics[width=0.95\linewidth]{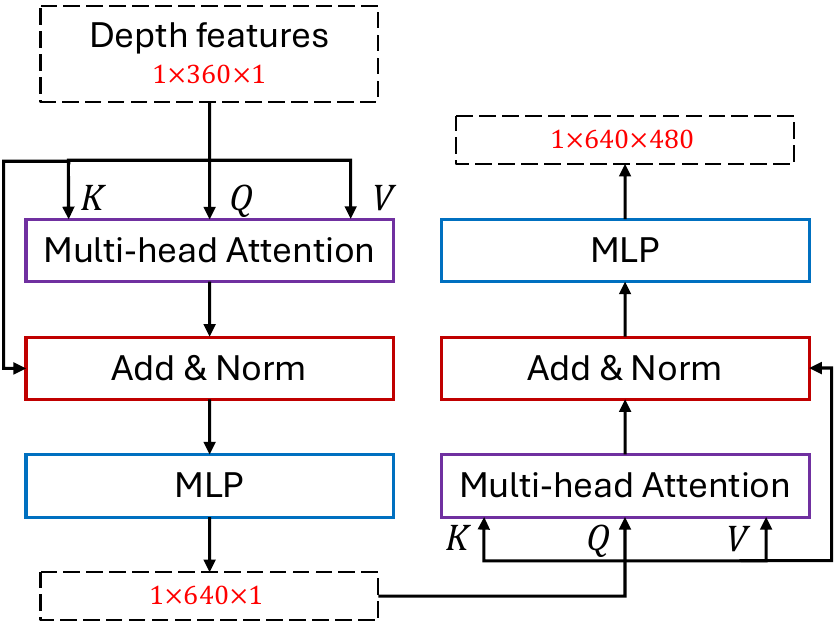}
\caption{Two-stage attention depth backbone. It transforms 1D depth vectors into spatially aligned embeddings by applying horizontal attention (beam alignment) followed by vertical attention (height projection), producing refined depth features for multi-modal fusion.}
\label{fig:depth_arch}
\end{figure}

\section{METHODOLOGY}
\subsection{Problem Formulation}
We tackle the RGB-D traversability segmentation problem within the few-shot segmentation (FSS) framework, aiming to meta-learn a multi-modal model that can quickly adapt a pretrained segmentation model to unseen indoor scenarios given limited labeled examples. Following recent advances in FSS \cite{wang2019panet,lu2021cwtfss,lang2022bam}, we adopt an episodic learning protocol for training and evaluation. The training set $\mathcal{D}_{train}$ comprises manually labeled RGB-D pairs, while the test set $\mathcal{D}_{test}$ is drawn from a large collection of unlabeled RGB-D pairs. To generate label masks for $\mathcal{D}_{test}$, we first train a fully supervised segmentation model on $\mathcal{D}_{train}$, achieving a mean Intersection over Union (mIoU) of 98.5\% under the 4-fold cross-validation. This high performance indicates its reliability for producing accurate masks used in evaluation. The test set $\mathcal{D}_{test}$ is then inferred on this fully supervised model to produce label masks for later evaluation.

\par
In each training episode, we sample $K$-shot examples from $\mathcal{D}_{train}$ to form the support set $S = \{S_{RGB}, S_D, S_{mask}\}$, where $S_{RGB}$ and $S_D$ denote the RGB image and depth map, respectively, and $S_{mask}$ is the corresponding ground-truth mask. Similarly, the query set $Q = \{Q_{RGB}, Q_D, Q_{mask}\}$ is drawn from $\mathcal{D}_{test}$. The model adapts to the support set to learn traversable indoor freespaces and infers segmentation masks for the query set. Given the large size of the unlabeled dataset (91,951 RGB-D pairs), 10,000 RGB-D pairs are randomly sampled for evaluation to ensure computational efficiency without compromising statistical validity.

\subsection{Methodology Overview}
Our proposed model is designed to efficiently extract and fuse heterogeneous information from RGB and depth modalities while preserving a lightweight training strategy. As illustrated in Figure \ref{fig:main_arch}, the model first embeds RGB and depth inputs into latent representations, denoted as $\{s_{RGB}, q_{RGB}\}$ and $\{s_{D}, q_{D}\}$, using modality-specific backbones (see Section \ref{sec:backbone}). These modality-specific features are then integrated through a multi-modal fusion module, yielding unified support and query features, $s$ and $q$, respectively. The fusion module is flexible in design and can incorporate state-of-art approaches such as \cite{zhang2023cmx,yin2023dformer}.  
\par
Next, the support feature $s$ is mask-pooled by the support mask $s_{mask}$ to generate positive ($s^{+}$) and negative ($s^{-}$) prototypes, corresponding to traversable freespace and obstacles. The query feature $q$ is then refined via cosine similarity with $s^{+}$ and $s^{-}$, producing positive ($q^{+}$) and negative ($q^{-}$) query features (see Section \ref{sec:contrast_att}). Finally, these query features are concatenated and passed to a lightweight decoder, which outputs the final query segmentation mask $q_{mask}$.

\subsection{RGB and Depth Backbones}
\label{sec:backbone}
The input tuple, consisting of an RGB image and a 1D depth vector, is first embedded by the RGB and depth backbones in parallel. Since our custom-collected RGB images have a resolution of $640 \times 480$, we design a lightweight RGB backbone composed of two convolutional layers with $3 \times 3$ kernels and a stride of 2, with GeLU activations \cite{hendrycks2016gaussian}.  

\par
To accommodate the format of our depth data—a single-row vector of dimension $[360]$ encoding both distance and viewing angle, we design a novel two-stage attention module for depth embedding, as illustrated in Figure \ref{fig:depth_arch}. The robot’s laser scanner captures measurements across a horizontal plane, where each pixel corresponds to a beam in the paired image and encodes the distance to obstacles. To avoid explicitly registering each depth value with its corresponding image pixel, the first stage of our depth module applies horizontal self-attention to learn embeddings $h_{d1}$ that align with the beams of the RGB image, as formulated in Eq. \ref{eq:depth_module1}.  

\begin{align}
q_{d} &= FC(x_{d}), \quad k_{d} = FC(x_{d}), \quad v_{d} = FC(x_{d}), \nonumber \\
h_{d1} &= \text{softmax}\left(\frac{q_{d} \cdot k_{d}^{T}}{\sqrt{m}}\right) \cdot v_{d},
\label{eq:depth_module1}    
\end{align}
where the input depth vector $x_d$ is linearly projected into \textit{Query}, \textit{Key}, and \textit{Value} representations in a self-attention block. The attended feature $h_{d1}$ captures horizontal depth information, where $m$ is the dimension of $x_d$. This stage also performs depth-to-image registration implicitly.
\par
The second stage vertically attends to the output of the first stage to produce a depth map $h_{d2}$ that matches the image height ($[480]$), as shown in Eq. \ref{eq:depth_module2}.

\begin{align}
h_{d1}' &= reshape(h_{d1}), \nonumber \\
q_{d2} &= FC(h_{d1}'), k_{d2} = FC(h_{d1}'), v_{d2} = FC(h_{d1}'), \nonumber \\
h_{d2} &= \text{softmax}\left(\frac{q_{d2} \cdot k_{d2}^{T}}{\sqrt{m_2}}\right) \cdot v_{d2},
\label{eq:depth_module2}    
\end{align}
where $m_2$ denotes the dimension of the output from the previous stage ($h_{d1}$). The resulting features are refined through a standard attention block, consisting of a residual connection with layer normalization and a feed-forward MLP. In this way, our proposed depth module effectively captures the geometric feature both horizontally and vertically and exploits the structural information contained in the 1D depth vector.

\par
Finally, the extracted RGB and depth features are fused using existing multi-modality fusion frameworks such as \cite{zhang2023cmx,yin2023dformer}, as defined in Eq. \ref{eq:fusion}.
\begin{align}
s &= Fusion(s_{RGB},s_D), \nonumber \\
q &= Fusion(q_{RGB}, q_D).
\label{eq:fusion}
\end{align}

\subsection{Contrastive Few-shot Learning}
\label{sec:contrast_att}
Most existing FSS approaches predict the query mask using only positive prototype matching \cite{snell2017prototypical,wang2019panet,lu2021cwtfss}, often overlooking the informative role of negative prototypes. A recent method \cite{lang2022bam} incorporates negative prototypes, but in a parametric manner that introduces additional trainable weights and increases model complexity. To address this limitation, we propose a novel contrastive few-shot learning strategy that fully exploits both positive and negative support prototypes in a non-parametric framework.
\par
Our method comprises two complementary branches. The first branch, positive-to-prototype ($p2p$), follows the traditional prototype matching paradigm \cite{wang2019panet}: the query’s positive feature $q^{+}$ is obtained by computing the cosine similarity between the query feature $q$ and the positive support prototype $s^{+}$. Query pixels with high similarity to $s^{+}$ are assigned to the traversable foreground class. 
\par
The second branch, negative-to-prototype ($n2p$), computes the cosine similarity between the query feature and the negative support prototype $s^{-}$, producing a negative feature representation $q^{-}$. The positive ($q^+$) and negative ($q^-$) features are then concatenated and passed to the decoder to predict the final segmentation mask. This contrastive design explicitly models both traversable and non-traversable cues, while introducing no additional parametric overhead.

\subsubsection{Positive Prototype Matching Branch}
The positive prototype matching branch ($p2p$) learns traversable freespaces by following the prototypical network paradigm in few-shot learning \cite{shaban2017oneshot}. Specifically, it identifies query pixels in $q$ that are most similar to the support set’s positive prototype $s^{+}$, yielding the positive query feature $q^{+}$.
\par
The support mask $s_{mask}$ mask-pools the support feature $s$, producing positive prototypes $s^{+}$ (freespace) and negative prototypes $s^{-}$ (obstacles), where the mask polarities $+$ and $-$ denote foreground and background, respectively. Mask-pooling, adapted from RoIAlign \cite{he2017maskrcnn}, outperforms traditional global average pooling (GAP) by preserving richer spatial information from the prototype feature map \cite{an2024few}.
\par
Finally, the positive query feature $q^{+}$ is derived by computing pixel-level cosine similarity between $q$ and $s^{+}$, as formulated in Eq. \ref{eq:p2p}:

\begin{align}
s^{+}&=mask\_pool(s_{mask}^{+},s), \nonumber \\
q^{+}&=cosine(s^{+},q).
\label{eq:p2p}
\end{align}

\subsubsection{Negative Contrastive Learning Branch}
To address the limited generalization ability of conventional positive prototype matching, we propose a Negative Contrastive Learning (NCL) branch ($n2p$). Relying solely on the $p2p$ branch often leads to overfitting to the support set, making it difficult to generalize across diverse freespace appearances (e.g., adapting from dark carpet to white ceramic tiles or colorful plastic floors). This issue is further exacerbated in few-shot settings due to the scarcity of training samples \cite{zhang2022mask,an2024few}.
\par
The $n2p$ branch mitigates this problem by leveraging negative prototypes in a non-parametric manner. Similar to the $p2p$ branch, negative prototypes $s^{-}$ are derived by $s_{mask}$ mask-pooling on the support feature $s$. We then compute the cosine similarity between $s^{-}$ and the query feature $q$, yielding the negative query feature $q^{-}$, which highlights pixels in $q$ resembling obstacles, as shown in Eq. \ref{eq:n2p}:

\begin{align}
s^{-}&=mask\_pool(s_{mask}^{-},s), \nonumber \\
q^{-}&=cosine(s^{-},q).
\label{eq:n2p}
\end{align}
\par
We adopt a non-parametric design to keep the $n2p$ branch as generic as possible. Introducing additional learnable layers risks overfitting, since these layers are randomly initialized and trained only on the limited support set in the FSS protocol. Our experiments confirm this issue, consistent with findings in \cite{lang2022bam}.

\par
Finally, the decoder concatenates both positive ($q^{+}$) and negative ($q^{-}$) query features to produce the final segmentation mask $q_{mask}$, as formulated in Eq.~\ref{eq:decode}: 
\begin{align}
q_{mask}&=decoder(q^{+},q^{-}).
\label{eq:decode}
\end{align}
Note that we omit explicit parametric fusing $q^+$ and $q^-$ like \cite{an2024few}, since the decoder already incorporates MLP layers (e.g., as in modern segmentation models such as \cite{yin2023dformer}), making additional ones unnecessary. 

\subsection{Other Training Details}
We highlight additional training details, given that our RGB-D pipeline incorporates both novel network modules and a new training strategy. To keep the meta-learning process lightweight and minimize learnable parameters, we only update the two-stage depth module and the decoder: the depth module is newly introduced, and the decoder is essential for any segmentation tasks. All other components, including the RGB backbone and fusion blocks, remain frozen. Notably, the mask-pooling and cosine similarity operations used in our training strategy do not introduce any additional learnable parameters.
\par
In summary, our methodology combines lightweight RGB-D backbones, non-parametric contrastive few-shot learning, and flexible multi-modal fusion to achieve robust traversability segmentation under few-shot conditions, while minimizing additional learnable parameters.
\begin{figure}[ht]
\centering
\includegraphics[width=0.47\linewidth]{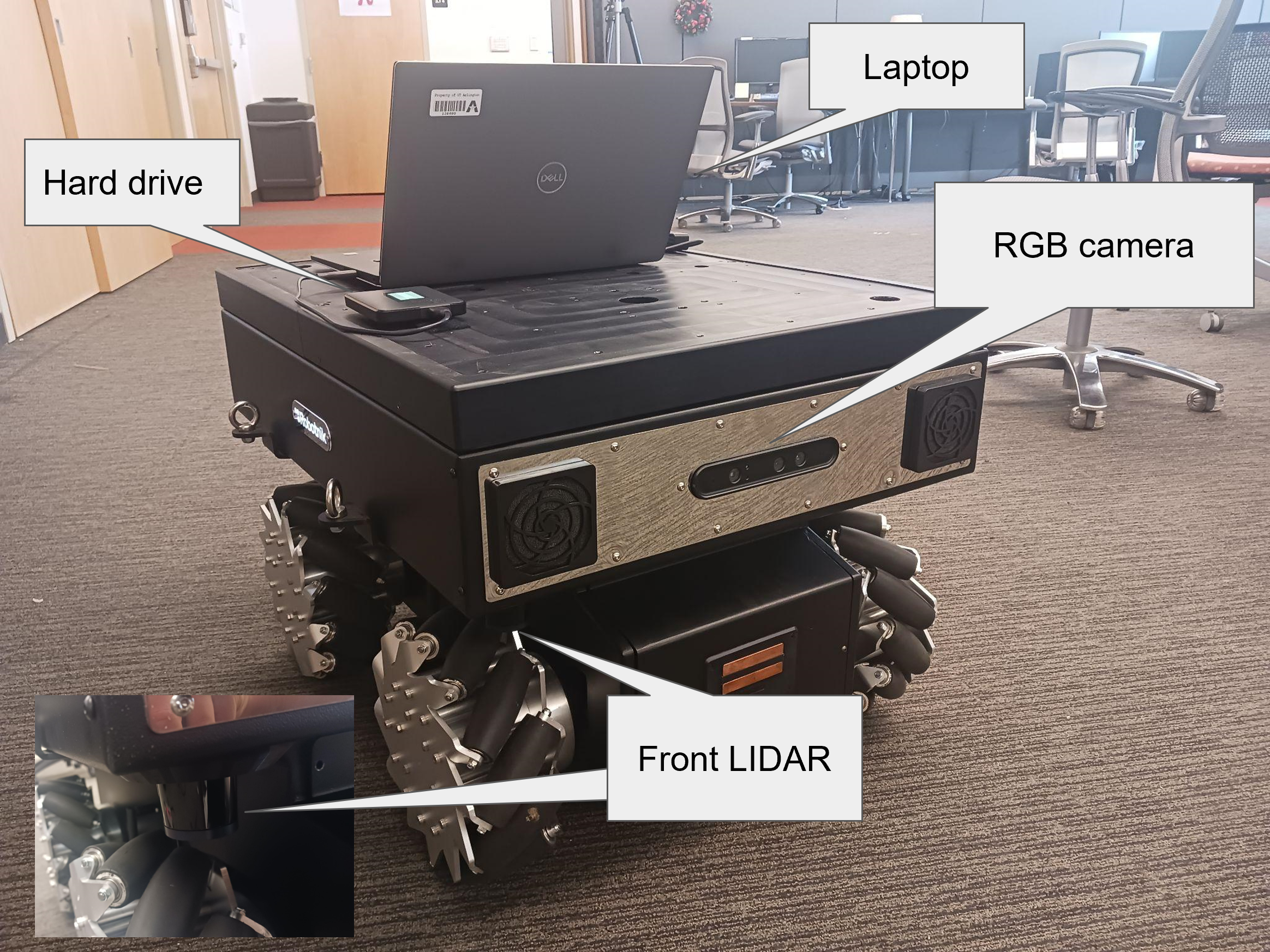}
\includegraphics[width=0.49\linewidth]{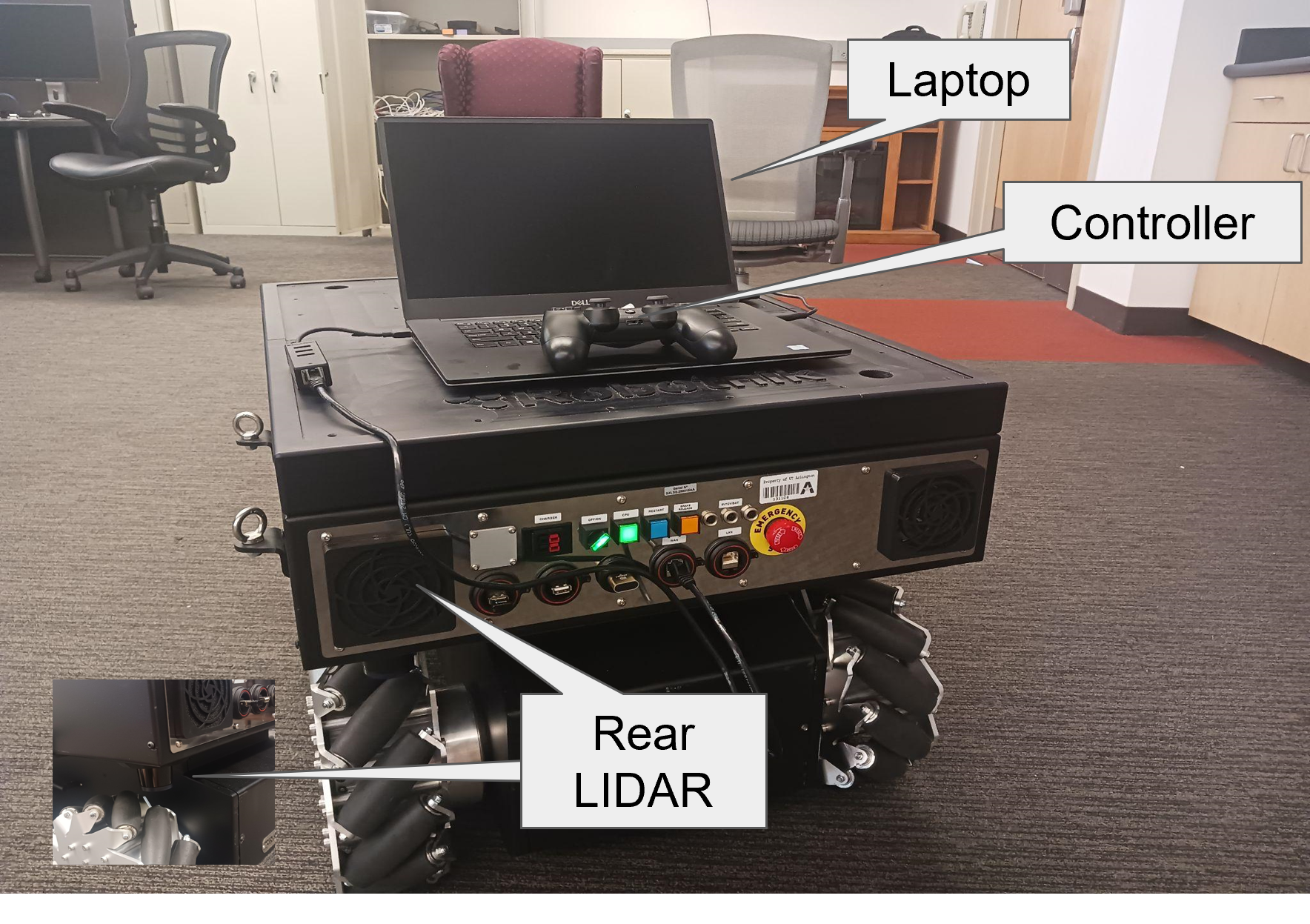}
\caption{Summit-XL Steel platform}
\label{fig:robot}
\end{figure}




\section{EXPERIMENTS}
\subsection{Dataset Collection}
We constructed an indoor traversability segmentation dataset using a tele-operated Summit-XL Steel robotic platform\footnote{\url{https://robots.ros.org/summit-xl-steel/}}, as shown in Figure \ref{fig:robot}. Multi-modal datastreams were recorded through ROS Melodic\footnote{\url{https://wiki.ros.org/melodic}} with the \textit{message\_filters} library\footnote{\url{http://wiki.ros.org/messagefilters}} to ensure synchronization between RGB images and depth laser data.
\par
The dataset covers multiple campus buildings, including classrooms, cafeterias, corridors, offices, and laboratories, providing diverse indoor layouts, lighting environments and floor materials. In total, the dataset contains 91,951 paired RGB and depth samples, of which 2,553 are manually annotated with freespace masks. Pseudo-labels for the unlabeled portion were generated by a strong fully-supervised teacher model achieving 98.5\% mIoU in 4-fold cross-validation, providing high-confidence reference masks. Similar practices can be found in \cite{wang2022semi}.
\par
Table~\ref{tab:dataset_stats} summarizes the dataset statistics across different buildings / domains. This multi-domain composition ensures a variety of visual and geometric conditions, making the dataset well-suited for evaluating generalization in RGB-D few-shot segmentation.

\begin{table}[ht]
\centering
\begin{tabular}{c|c|c|c|c}
\toprule
Building & ELB & ERB & NH & UC \\
\midrule
RGB-D pairs & 8732 & 5590 & 4610 & 36072 \\
Manually labeled & 649 & 658 & 433 & 521 \\
\midrule
Building & WH & Mocap & Heracleia & Total \\
\midrule
RGB-D pairs & 10899 & 10872 & 17295 & 91951 \\
Manually labeled & 292 & - & - & 2553 \\
\bottomrule
\end{tabular}
\caption{Statistics of the custom-collected indoor RGB-D traversability dataset.}
\label{tab:dataset_stats}
\end{table}

\subsection{Metrics and Implementation Details}
We report performance using mean Intersection-over-Union (mIoU), the standard metric for semantic segmentation, and adopt an episodic training protocol consistent with prior work \cite{lu2021cwtfss}. Regarding trainable parameters, the RGB backbone and the multi-modality fusion module from DFormer are frozen \cite{yin2023dformer}, while updating only the learnable modules: the proposed depth backbone and the final decoder. The RGB backbone is pretrained on ImageNet-1K \cite{russakovsky2015imagenet}, and the fusion module is pretrained on NYUDepthv2 \cite{silberman2012nyuv2}.
\par
Our dataset is organized into episodes, each consisting of a support set and a query set, where every sample includes an RGB image paired with its corresponding 1D depth vector. We evaluate under both 1-shot and 5-shot configurations, following standard FSS protocols \cite{lu2021cwtfss,zhang2022mask,li2021adaptive}. For each episode, the trainable modules are adapted on the support set for 120 epochs using cross-entropy loss and the AdamW optimizer \cite{kingma2014adamw} with an initial learning rate of $6\times10^{-5}$ and scheduled by WarmUpPolyLR with a power of 0.9, a weight decay of 0.01, and 5 warm-up epochs.

\begin{table*}[ht]
\begin{center}
\begin{tabular}{c|c|ccc|ccc|c}
\toprule
\multirow{2}{*}{Backbone} & \multirow{2}{*}{Method} & \multicolumn{3}{c|}{1-shot} & \multicolumn{3}{c|}{5-shot} & \multirow{2}{*}{Trainable / Total Params} \\[1pt]
\cline{3-8}
\noalign{\vskip 2pt}
& & Freespace & Obstacles & mIoU & Freespace & Obstacles & mIoU & \\
\midrule
\multirow{4}{*}{CMX \cite{zhang2023cmx}} & PANet \cite{wang2019panet} & 74.8 & 52.01 & 63.4 & 76.24 & 54.81 & 65.52 & 2.5M / 65.7M\\
& CWT \cite{lu2021cwtfss} & 83.51 & 64.82 & 74.16 & 84.87 & 64.38 & 74.68 & 1.2M / 98M \\
& BAM \cite{lang2022bam} & 86.03 & 73.39 & 78.91 & 85.67 & 74.69 & 80.19 & 2.8M / 102M\\[1pt]
\cline{2-9}
\noalign{\vskip 2pt}
& NCL (ours) & 91.5 & 82.55 & 87.03 & 91.84 & 83.94 & 87.91 & 4.6M / 59.7M \\
\midrule
\multirow{4}{*}{DFormer \cite{yin2023dformer}} & PANet & 76.45 & 53.16 & 64.8 & 78.54 & 56.46 & 67.5 & 4.3M / 14.7M \\
& CWT & 85.35 & 66.25 & 75.8 & 87.43 & 66.8 & 77.11 & 1.2M / 46.7M \\
& BAM & 87.93 & 75.01 & 81.47 & 88.26 & 76.94 & 82.61 & 4.9M / 51.6M\\[1pt]
\cline{2-9}
\noalign{\vskip 2pt}
& NCL (ours) & 93.52 & 84.37 & 88.95 & 94.61 & 86.5 & 90.56 & 4.4M / 29.6M \\
\bottomrule
\end{tabular}
\end{center}
\caption{Quantitative results on the indoor traversability dataset under 1-shot and 5-shot settings. Our proposed NCL achieves the best mIoU across all settings while requiring only a small fraction of trainable parameters, since most backbone weights are frozen during adaptation.}
\label{tab:mm_quan}
\end{table*}

\subsection{Quantitative Results}
Table \ref{tab:mm_quan} presents the traversability segmentation results on our indoor dataset. To enable comparing with RGB-D segmentation baselines that require 2D depth maps directly, we convert our 1D depth vector into a pseudo-2D map by warping to image width and duplicating along the image's height.
\par
Our method, NCL, consistently outperforms state-of-the-art few-shot segmentation methods across both 1-shot and 5-shot settings. On average, NCL improves mIoU by 5–15 points compared to PANet, CWT, and BAM, with the most significant gains observed in 1-shot scenarios (8 points more than BAM). These improvements are consistent across two fusion backbones, CMX and DFormer, demonstrating the robustness and generality of our approach. All reported metrics are means over 5 independent runs with different random seeds for episodic sampling; standard deviations were consistently $<0.4\%$ in mIoU and are omitted for brevity given the large test set size.
\par
The performance gains can be attributed to key design choices. PANet, a traditional prototype-matching method, overlooks negative prototypes and employs global average pooling, thereby losing discriminative information from the support set. CWT transfers only the classifier layer of a transformer-based backbone, limiting its ability to model complex cross-modal interactions. BAM attempts to separate background objects but relies on additional parametric layers, which are prone to overfit under few-shot conditions. In contrast, NCL leverages non-parametric negative contrastive cues, capturing both positive and negative information without extra trainable parameters, resulting in more robust generalization to unseen query scenarios.
\par
These results indicate that explicitly modeling negative prototypes via contrastive learning significantly improves few-shot RGB-D traversability segmentation, making our approach both accurate and backbone-agnostic.

\begin{figure*}[ht]
\centering
\begin{subfigure}{0.98\textwidth}
\includegraphics[width=\linewidth]{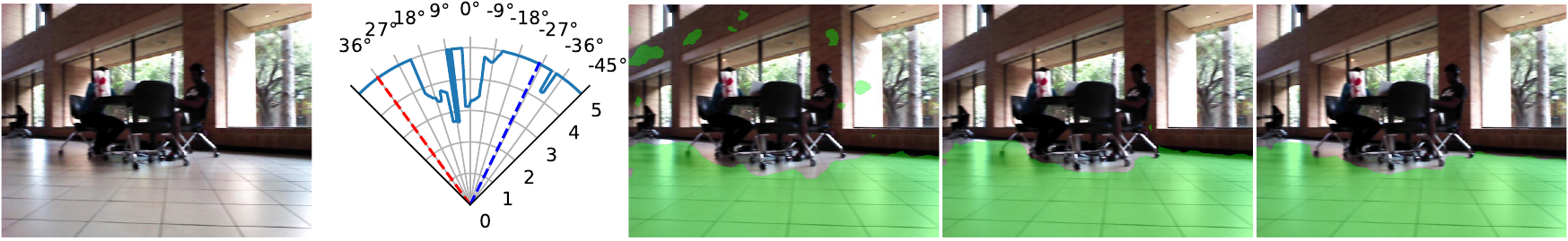}
\end{subfigure}
\begin{subfigure}{0.98\textwidth}
\includegraphics[width=\linewidth]{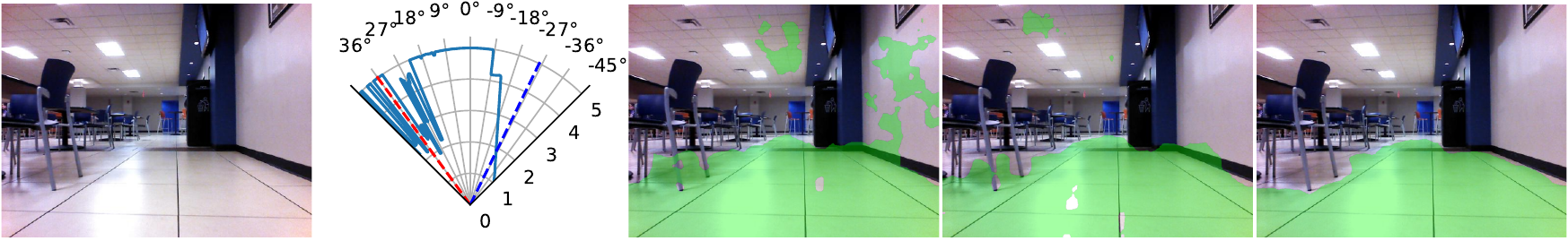}
\end{subfigure}
\begin{subfigure}{0.98\textwidth}
\includegraphics[width=\linewidth]{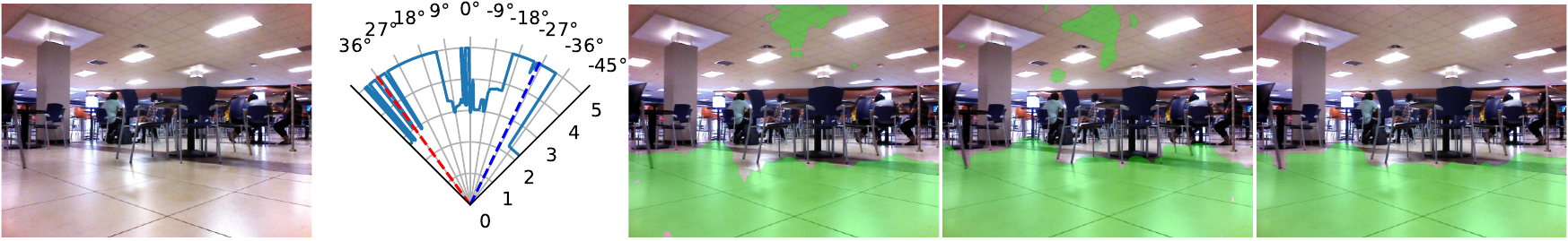}
\end{subfigure}
\begin{subfigure}{0.98\textwidth}
\includegraphics[width=\linewidth]{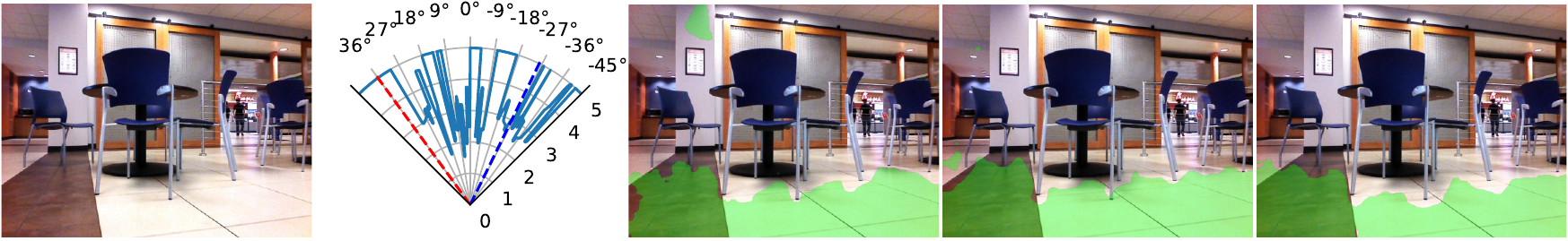}
\end{subfigure}
\caption{Qualitative results on indoor traversability segmentation. Each row shows (1) the query RGB image, (2) the corresponding depth vector, (3) predictions without the two-stage depth attention module and without NCL, (4) predictions with the depth module but without NCL, and (5) predictions from the full model. The proposed depth module helps separate floors from walls/ceilings, while the NCL branch further improves recognition of thin obstacles (e.g., chair legs).}
\label{fig:mm_qual1}
\end{figure*}

\subsection{Qualitative Results}
A primary objective of this work is to overcome the limitations of pure vision-based models in recognizing thin obstacles, such as chair legs. Although these objects often occupy less than 1\% of image pixels and have little impact on mIoU, they pose serious safety risks to autonomous agents and surrounding people. To highlight the effectiveness of our method, we present representative qualitative examples in Figure \ref{fig:mm_qual1}.
\par
Each row shows (1) the query RGB image, (2) the corresponding depth vector, where the left and right boundaries of RGB sights are shown in red and blue dashed lines, and values represents the distance to obstacles up to 5 meters, (3) predictions from a model without the two-stage depth attention module and without the NCL branch, (4) predictions with the depth module but without NCL, and (5) predictions from our complete model with both components enabled.
\par
We observe three trends:
\begin{enumerate}
\item Without the depth module (col. 3): the model frequently confuses floors with visually similar walls or ceilings, leading to significant false positives.
\item With the depth module only (col. 4): the model better separates floors from walls/ceilings, but still fails to exclude thin objects such as chair legs.
\item Full model with NCL (col. 5): the segmentation improves significantly, successfully excluding thin obstacles and yielding clean, safe freespace predictions.
\end{enumerate}
These results visually confirm the complementary benefits of the proposed two-stage depth attention module and the negative contrastive learning branch.

\begin{table}[ht]
\centering
\begin{tabular}{c|c|c|c|c}
\toprule
 & & \multicolumn{3}{c}{1-shot} \\[1pt]
\cline{3-5}
\noalign{\vskip 2pt}
Backbone & Setting & Freespace & Obstacles & mIoU \\
\midrule
\multirow{4}{*}{DFormer \cite{yin2023dformer}} & -$H$ -$W$ & 85.84 & 69.1& 77.47\\
& -$H$ +$W$ & 88.63& 71.44& 80.03\\
& +$H$ -$W$ & 90.32& 76.74& 83.53\\
& +$H$ +$W$ & 93.52& 84.37& 88.95\\
\bottomrule
\end{tabular}
\caption{Ablation study on the two-stage depth attention module, where $H$ and $W$ represent horizontal and vertical attention blocks respectively.}
\label{tab:ablation_depth}
\end{table}

\subsection{Ablation Study}
\subsubsection{Two-stage Attention Depth Module}
We conduct an ablation study on the proposed two-stage attention depth module using the DFormer backbone under the 1-shot FSS setting. Results are summarized in Table \ref{tab:ablation_depth}.
\par
The baseline setting (-$H$ -$W$) directly warps the depth vector to the image width and duplicates it across rows without any attention mechanism. This achieves the lowest performance (77.47 mIoU), as it fails to exploit the geometric cues inherent in the depth signal. Adding only width attention (-$H$ +$W$) improves mIoU by +2.6, since it better aligns depth values with horizontal pixel positions. Introducing only height attention (+$H$ -$W$) achieves a larger gain of +6.1 over the baseline, highlighting the importance of vertical depth cues. For instance, a sudden decrease in depth often corresponds to nearby obstacles, directly constraining the freespace mask along the image height. 
\par
Finally, the full two-stage attention module (+$H$ +$W$) delivers the best performance at 88.95 mIoU, a +11.5 improvement over the baseline. This demonstrates that horizontal and vertical attentions provide complementary benefits, and their joint modeling is crucial for capturing fine-grained geometric structures. While the height attention effectively propagates learned structural priors from RGB vertically, it can introduce hallucinations in regions with insufficient cues - as illustrated in our attached demo video.

\begin{table}[ht]
\centering
\begin{tabular}{c|c|c|c|c}
\toprule
 & & \multicolumn{3}{c}{1-shot} \\[1pt]
\cline{3-5}
\noalign{\vskip 2pt}
Backbone & Setting & Freespace & Obstacles & mIoU \\
\midrule
\multirow{2}{*}{DFormer \cite{yin2023dformer}} & +$p2p$ -$n2p$ & 88.36 & 72.96& 80.66\\
& +$p2p$ +$n2p$ & 93.52& 84.37& 88.95\\
\bottomrule
\end{tabular}
\caption{Ablation study on the negative contrastive learning branch $n2p$, where $p2p$ represents the traditional positive prototype matching branch.}
\label{tab:ablation_n2p}
\end{table}

\subsubsection{Negative Contrastive Learning Branch}
To assess the contribution of the proposed negative contrastive learning (NCL) branch, we ablate the $p2p$ and $n2p$ branches under the 1-shot setting with the DFormer fusion backbone. Results are shown in Table \ref{tab:ablation_n2p}.
\par
Using only the positive prototype matching branch (+$p2p$ -$n2p$) achieves 80.66 mIoU. Adding the NCL branch (+$p2p$ +$n2p$) improves performance gain of +8.3. Notably, the largest improvement is observed in the obstacle class (+11.4 IoU), compared to +5.2 for freespace. This confirms that explicitly modeling negative prototypes significantly enhances the model’s ability to separate obstacles from freespace—an aspect that purely positive prototype matching tends to overlook.
\par
Therefore, the $n2p$ branch not only improves overall segmentation accuracy but also addresses a critical weakness of conventional FSS methods, i.e., the inability to robustly reject non-traversable regions.
\section{Conclusion}
We introduce a multi-modal RGB-D few-shot segmentation framework to improve indoor traversability analysis, with a particular focus on recognizing and excluding thin obstacles that are often overlooked by vision-only models. By digesting RGB images alongside 1D laser depth data, our approach leverages complementary geometric cues while remaining lightweight in terms of trainable parameters. To address the limited availability of labeled data and the need for generalization to unseen environments, we adopted a few-shot learning paradigm and proposed a novel negative contrastive learning branch to complement traditional positive prototype matching. Extensive experiments demonstrate that our framework significantly improves both qualitative and quantitative performance, particularly in challenging scenarios with thin obstacles. We believe this work provides a promising direction for safer and more robust indoor robotic navigation, and we plan to release our dataset and code to further support research in the robotics and assistive systems community.

\section*{Acknowledgements}
We thank Christos Sevastopoulos and Sneh Acharya for their invaluable assistance in collecting and annotating the dataset.

\bibliographystyle{IEEEtran.bst}
\bibliography{root.bib}

\end{document}